\documentclass{bmvc2k}

\usepackage{graphicx}
\usepackage{adjustbox}
\usepackage{booktabs}
\usepackage{sidecap}
\usepackage{floatrow}

\title{Video Time: Properties, Encoders and Evaluation}

\addauthor{Amir Ghodrati \\ Efstratios Gavves \\ Cees G.~M. Snoek}{{a.ghodrati, egavves, cgmsnoek }@uva.nl}{1}

\addinstitution{
 QUVA Lab\\
 University of Amsterdam\\
 Netherlands
}

\runninghead{Ghodrati, Gavves, Snoek}{Video Time}

\def\eg{\emph{e.g}\bmvaOneDot}

\def\etal{\emph{et al}\bmvaOneDot}
\def\ie{\emph{i.e}\bmvaOneDot}

\newcommand{\ourmodel}{Time-Aligned DenseNet\xspace}

\newcommand{\tsm}{Sequential Video Time Encoders\xspace} 
\newcommand{\thm}{Hierarchical Video Time Encoders\xspace} 
\begin{document}

\maketitle

\begin{abstract}
Time-aware encoding of frame sequences in a video is a fundamental problem in video understanding. While many attempted to model time in videos, an explicit study on quantifying video time is missing. To fill this lacuna, we aim to evaluate video time explicitly. We describe three properties of video time, namely \emph{a)} temporal asymmetry, \emph{b)} temporal continuity and \emph{c)} temporal causality. Based on each we formulate a task able to quantify the associated property. This allows assessing the effectiveness of modern video encoders, like C3D and LSTM, in their ability to model time. Our analysis provides insights about existing encoders while also leading us to propose a new video time encoder, which is better suited for the video time recognition tasks than C3D and LSTM. We believe the proposed meta-analysis can provide a reasonable baseline to assess video time encoders on equal grounds on a set of temporal-aware tasks\footnote{Code is available at \href{https://aghodrati.github.io/videotime.html}{https://aghodrati.github.io/videotime.html}}. 
\end{abstract}
\section{Introduction} \label{sec:intro}
The goal of this paper is to investigate and evaluate the manifestation of time in video data. Modeling time in video is crucial for understanding \cite{fernando2015modeling,wang2016temporal}, prediction \cite{ryoo2011human,de2016online} and reasoning \cite{goyal2017something,damen2018epic}. Hence, it received increased attention in the computer vision community lately. Often in the form of advanced video encoders for activity recognition, \eg \cite{simonyan2014two,donahue2015long,sigurdsson2017asynchronous,tran2015learning,carreira2017quo}. However, it is well known that actions may also be recognized by (static) appearance cues in the scene~\cite{ikizler2010object}, or from characteristic objects \cite{JainCVPR15}. Hence, it is hard to assess the contribution of modeling time using this task. Rather than evaluating time implicitly as part of proxy tasks like activity recognition, we prefer an explicit analysis of video time.

We are encouraged by two recent meta-analysis studies that assess video time as well. In \cite{sigurdsson2017actions} Sigurdsson \etal examine datasets, evaluation metrics, algorithms, and potential future directions in the context of action classification. They find that video encoders have to develop a temporal understanding to differentiate temporally similar but semantically different video snippets. Very recently Huang \etal \cite{huangmakes2018} measured the effect of reducing motion in C3D video encoders during an ablation analysis on action classification datasets. Instead of removing motion in an action classification task, we prefer to evaluate the effect of video time in encoders like C3D, LSTM and our own proposal, on three time-aware tasks.

This paper makes three contributions to the meta-analysis of time in video. Inspired by three properties of video time, we first present three time-aware video recognition tasks allowing to evaluate the temporal modeling abilities of video encoders.
Second, we categorize video encoders into two general families.
From their analysis, we derive a new video time encoder, specifically designed to capture the temporal characteristics of video. Third, we evaluate our video encoders, C3D and LSTM with respect to their temporal modeling abilities on the three time-aware video recognition tasks and discuss our findings.

\section{Related Work} \label{sec:related}
Time in video is an ambiguous concept which is hard to be quantified.~\cite{pickup2014seeing,arrow18} strive to observe the time signal within video frames by distinguishing natural ordered frames from reversed frames. Isola~\etal~\cite{isola2015discovering} see this ``arrow of time'' by discovering the transformations that objects undergo over time. Zhou~\etal~\cite{zhou2015temporal} propose the task of predicting the correct temporal ordering of two video snippets. We also consider arrow of time prediction to quantify the temporal asymmetry, but rather than aiming to find the best way to solve the task, we evaluate how capable modern video encoders are in addressing this task.

%

We also focus on predicting future frames, as being a task that directly connects to temporal continuity.
Recently, predicting the future has received significant attention in computer vision.
Different forms of future prediction have been proposed, for activities \cite{ryoo2011human,hoai2014max,de2016online,zeng2017visual}, 
human trajectories~\cite{pintea2014deja,walker2016uncertain},
body poses~\cite{fragkiadaki2015recurrent,walker2017pose}, 
visual representations~\cite{vondrick2016anticipating},
or even to generate the pixel-level reconstructions of future frames~\cite{ranzato2014video,xue2016visual,mathieu2015deep,vondrick2017generating,lotter2016deep,villegas2017learning}.
%
However, generating future pixels is ill-posed: there exist infinite possible futures and generating pixels is a separate machine learning challenge~\cite{kingma2013auto,gregor2015draw, goodfellow2014generative}.
We, therefore, recast the future frame prediction task as a forward-frame retrieval problem where the goal is to select the correct future frame out of $C$ possible choices.
Clearly, casting future frame prediction as retrieval cannot be deployed in practice, since in practice we cannot access future frames.
However, it allows for a clear, well-understood and consistent evaluation framework. 

Most works for encoding temporal relations are evaluated on action recognition tasks~\cite{gaidon2013temporal,wang2013action,fernando2015modeling,simonyan2014two,donahue2015long,tran2015learning,wang2016temporal}. A video is encoded by learning either the frame order~\cite{fernando2015modeling}, short-term motion patterns~\cite{simonyan2014two}, frames dependencies~\cite{donahue2015long,srivastava2015unsupervised,sigurdsson2017asynchronous}, spatio-temporal representation~\cite{tran2015learning,ji20133d,carreira2017quo} or fully connected layers~\cite{zhou2017temporal}.
However, often, recognizing an action can be attributed to several factors, besides temporal modeling, such as the scene type, the appearance of the actors, particular poses, and so on.
%
Alternatively, actions can be defined procedurally, as an ordered set of sub-actions.
The advantage is that procedures effectively determine a cause and effect, and, therefore, are temporally causal.
For instance,~\cite{goyal2017something} defines actions as templates like \emph{taking something from somewhere} or \emph{putting something next to something}. In such a setting, temporal modeling of detailed sub-actions is required for successful recognition.
Thus, we propose to evaluate temporal causality on action templates instead of standard action classes.

\section{Properties of Video Time} \label{sec:tasks}

In this section, we describe three properties of video time and based on each, we formulate a task able to quantify the associated property.

\begin{figure}[t]
\includegraphics[width=0.9\linewidth]{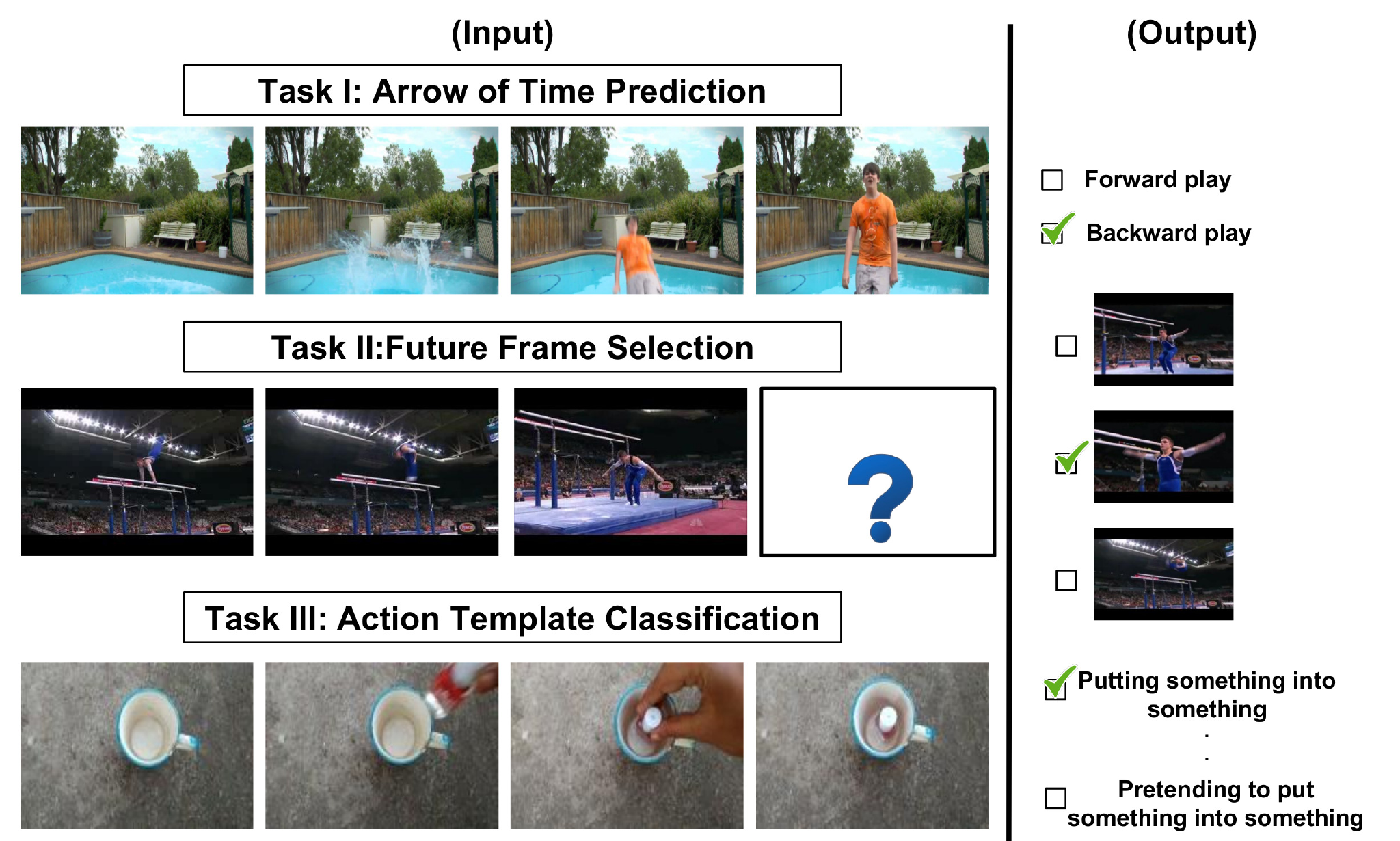}
\caption{Inspired by three properties of time, detailed in Section~\ref{sec:tasks}, we present three time-aware video tasks that allow to evaluate video encoders in light of their temporal modeling abilities.}
%
\label{fig:tasks}
\end{figure}

\noindent\newline\textbf{Property I: Temporal Asymmetry.}
%
%
A first important property of time is its asymmetric nature,
%
%
namely the fact that there is a clear distinction between the forward and the backward arrow of time.
A question, therefore, is to what extent time asymmetry is also observable and quantifiable by video encoders.
%
%
Similarly to Pickup \etal~\cite{pickup2014seeing}, we adopt the \emph{\textbf{arrow of time prediction}} task, a binary classification task, which aims to predict whether a video is played forwards or backwards.
Different from~\cite{pickup2014seeing}, we focus on exploring how capable state-of-the-art video encoders are in addressing this task, instead of finding time-sensitive low-level features to solve this task.


\noindent\newline\textbf{Property II: Temporal Continuity.}
%
A second important property of time is continuity.
Because of temporal continuity, future observations are expected to be a smooth continuation of past observations, up to a given temporal scale.
A time-aware video encoder should be able to predict future frames by its ability to model temporal continuity. In the second task we measure how capable video encoders are in predicting future frames. We introduce the task of \emph{\textbf{future frame selection}}, where we assume we have access to the correct future frame at $T-\tau$ seconds from the currently observed frame $x_\tau$. The goal is to predict the correct future frame among other incorrect ones.
By careful sampling of the incorrect future frames we can control the difficulty of the evaluation.
This setup allows expressing future frame prediction as a well-defined classification task with a well-defined evaluation procedure.

\noindent\newline\textbf{Property III: Temporal Causality.}
A third important aspect of time in videos relates to causality.
An interesting observation for the traditional action classification task in YouTube-based UCF101 dataset~\cite{soomro2012ucf101} is that it does not necessarily need causal reasoning, meaning that any permutation of the frame order will not change the action classification performance much~\cite{zhou2017temporal}.
For instance, recognizing \emph{playing basketball} is less determined by the precise order of the video frames, and mostly by the presence of the basketball court.  
However, a good video encoder should capture such causal relations.
%
To formulate this property, we focus on \emph{\textbf{action template classification}}. In this setup, classes are formulated as template-based textual descriptions~\cite{goyal2017something}, where an action is characterized by temporal relations of detailed intuitive physics. An example of such a template-based action class is \emph{Putting something into something}, which can only be successfully discriminated from \emph{Pretending to put something into something} if the temporal relations are accurately modeled.

We summarize the tasks able to evaluate the properties of video time in Fig~\ref{fig:tasks}.

\section{Video Time Encoders} \label{sec:models}
We are interested in a video encoder $f(x_1, \dots, x_t)$ capable of modeling the time signal present in frames $x_1, \dots, x_t$.
To structure our analysis, we first group existing video encoders into two general families namely \tsm and \thm, and then we describe our proposed video time encoder.

\noindent\newline\textbf{\tsm.}
This family of video encoders views temporal data as infinite time series $x_1, x_2, ..., x_t$, where each time step $t$ is associated to a state $h_t$.
Formally, in each step these encoders model the transition between preceding states to the current state by
$h_{t} \propto f(x_1, \dots , x_t, h_1, \dots , h_{t-1}; \theta),$
where $f(.)$ is a function parametrized by $\theta$. 
Recurrent neural networks and variants (LSTM~\cite{hochreiter1997long}, GRU~\cite{cho2014learning}), as well as Hidden Markov Models~\cite{baum1966statistical} all belong to the sequential family of video encoders.
Particularly, RNNs fit to the video domain due to their long-range temporal recursion.
In their simplest form, recurrent neural networks model the hidden state at time step $t$ by
\begin{equation}
h_t = f(x_t, h_{t-1}; \theta) = \sigma_h(W_h \cdot x_t + U_h \cdot h_{t-1} + b_h), 
\label{tsm_rnn}
\end{equation}
where $h_{t-1}$ summarizes all the information about the past behaviour of the frame sequence.
The function $\sigma_h$ is a non-linearity
and the parameters $\theta=\{W_{\cdot}, U_{\cdot}, b_{\cdot}\}$ are shared and learned with backpropagation through time.
As such, a single parameter set $\theta$ suffices to capture the temporal relations between subsequent variables $x_i$ and $x_{i+1}$. 
A special variant of RNNs are LSTMs, which have shown to be particularly strong in learning long-term dependencies and, thus, have been widely used in the vision community for modeling frame sequences~\cite{donahue2015long,srivastava2015unsupervised}. Hence, as a representative sequential encoder, we choose LSTMs in this work.

\noindent\newline\textbf{\thm.} 
This family of encoders views video as a hierarchy of temporal finite ``patches'' of horizon $T$, where each patch is defined as $H^{(l)} = [h_i^{(l)}, \dots , h_{i+T}^{(l)}]$.
Formally, in level $l$ of the hierarchy, $h^{(l)}$ is defined as
\begin{equation}
h^{(l)} \propto \sigma_h \Big( h^{(l-1)} * W^{(l)} + b^{(l)} \Big), 
\label{thm}
\end{equation}
where $h_i^{(0)}=x_i$ and $*$ denotes the convolution operator. 
This is similar to the way convolutional neural networks conceive an image as a structure of image patches.
Like convolutional neural networks, \thm define successive layers $(l)$ of non-linearities.
Specifically, at layer $(l)$ a set of temporal templates ($W^{(l)} , b^{(l)}$) (layer parameters) are learned, based on co-occurrence patterns present in the input.
Similar to convolutional neural networks for images, different layers have different parameters.


C3D~\cite{tran2015learning,ji20133d} and its variants (like i3D~\cite{carreira2017quo} and LTC~\cite{varol2017long}) belong to the temporal hierarchy family of models, where spatial and temporal filters are learned jointly using 3D convolutions. 
%
As a representative hierarchical encoder, we choose C3D in this paper, which has been widely used in the literature, \eg~\cite{tran2015learning,ji20133d}.

\noindent\newline\textbf{Discussion.}
The main difference between \tsm and \thm is in their handling of the temporal sequence. Sequential encoders, like LSTMs, view temporal data as time series, and embed all information from the past into a state variable. They then learn a transition function from the current to the next state. By construction, the model converges in the limit to a single, static transition matrix shared between all states in the chain. For video frame sequences, this transition matrix must then be able to model the temporal patterns in the input sequences.
Under the noisy and non-stationary reality of video content, learning the transition function is hard.
However, sequential encoders focus on learning transitions from one temporal state to the other. Thus they are able, in theory, to model conditional dependencies through time.

In contrast, hierarchical encoders rely on convolutions, which are equivalent to correlations after rotation by $\pi$.
Thus, hierarchical video time encoders focus on the correlations between input elements, rather than their conditional temporal dependencies.
Consequently, hierarchical encoders are well-suited either when temporal correlations suffice for the task, or when correlations coincide with causations for a given task.
By viewing sequences as ``finite temporal patches'' in temporal hierarchies, hierarchical encoders can learn specialized filters to recognize specific temporal patterns of different abstractions.
Because they introduce many more learnable filter parameters, however, they are more prone to overfitting.
%

\noindent\newline\textbf{Proposed \ourmodel.}
%
To overcome the limitations of sequential and hierarchical video encoders in time-aware video recognition tasks, we propose the \ourmodel, a neural architecture inspired by DenseNet~\cite{huang2017densely}.
Like DenseNet, it has densely connected patterns, \ie each layer is connected to all its preceding layers.
Unlike DenseNet, the layers are aligned along the temporal dimension rather than the spatial one.
Formally, we describe the \ourmodel by
\begin{eqnarray}
h_t &=& f(x_t, h_1, \dots h_{t-1}; \theta_t) .
\label{ours}
\end{eqnarray}
The proposed model can be seen as a hybrid between sequential and hierarchical encoders. 
Similar to RNNs, the \ourmodel also views video frames as time series.
Similar to C3D, and unlike RNNs, however, it does not share parameters through time.

Because of the non-shared parameterization \ourmodel enjoys several benefits.
First, the encoder has a greater flexibility in modeling temporal transitions compared to LSTM.
Second, there is no recursion and standard backpropagation suffices, instead of backpropagation through time.
Thus, the encoder is able to exploit all the modern deep learning advances for a better training.
This is important, as the shared parameterization can easily lead to chaotic behavior in RNNs~\cite{pascanu2013difficulty}, forcing RNNs to a more conservative optimization regime.
Last, \ourmodel has explicit access to \emph{all} previous hidden states, unlike RNNs that have only implicit access.

\begin{figure}[t]
\includegraphics[width=0.9\linewidth]{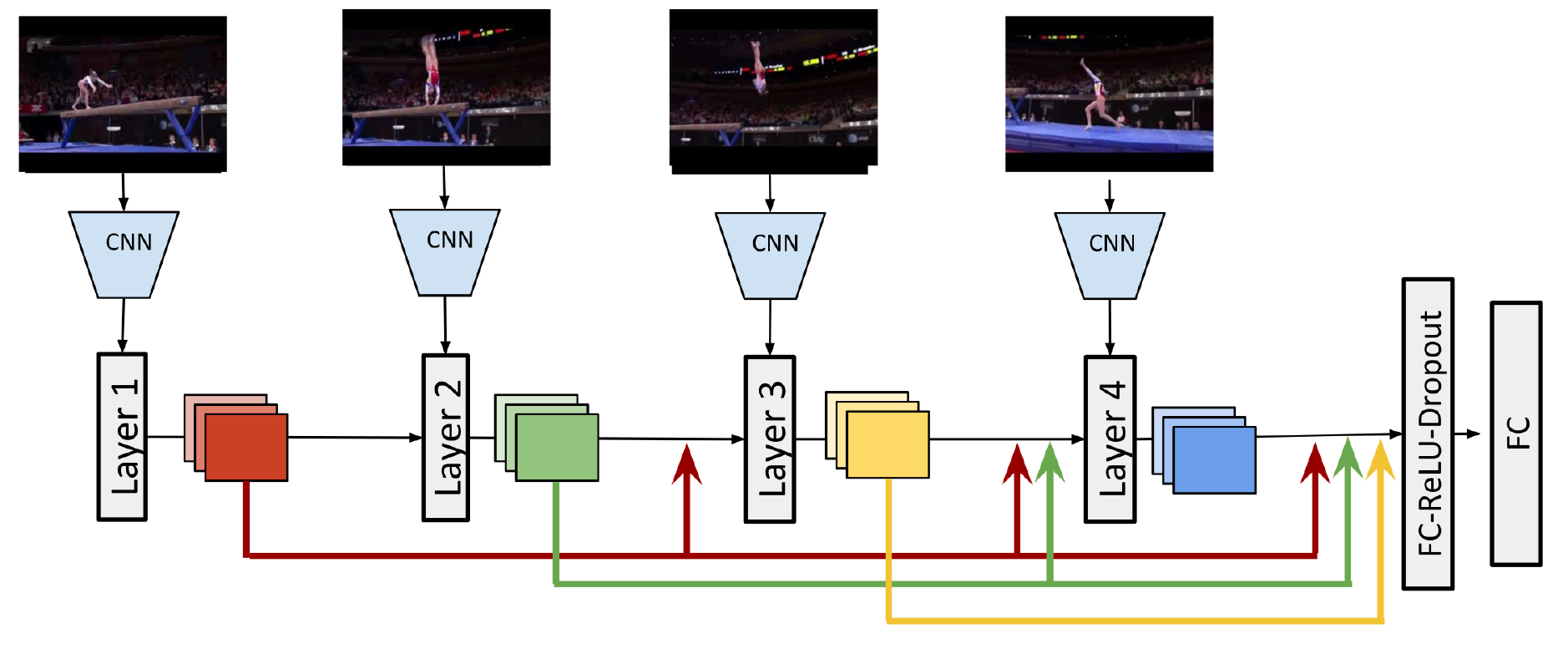}
\caption{Proposed \ourmodel architecture. Individual frames are encoded using a CNN and then fed to an encoder where each layer represents a time-step. Inspired by \cite{huang2017densely}, each layer is connected to all its preceding layers. At time step $T$, the encoder outputs $K \times T$ feature maps as representation of the video.}
\label{fig:model}
\end{figure}

The general architecture of the \ourmodel is shown in Fig.~\ref{fig:model}. 
Although we have no restrictions on how to represent the state variables $h_t$, since we focus on video we opt for convolutional feature maps.
We start from a sequence of individual frames from a video, encoded using a convolutional neural network, as LSTM.
For each time step in the video, we have an associated layer in the network. Specifically, we implement each layer per time step as a DenseNet block~\cite{huang2017densely}:
%
\begin{eqnarray}
f^a & = & \text{Concat}([h_1, ..., h_{t-1}, x_t]) \nonumber \\
h_t & = & \text{DropOut}(\text{Conv2D}(\text{ReLU}(\text{BatchNorm}(f^a, \theta_{t}^{bn})); \theta_{t}^{conv}, K)),
\end{eqnarray}
%
%
%
where $K$ is the number of output feature maps dedicated to layer $t$. It is worth noting that such a design makes our architecture fully convolutional. One can see the per time step feature maps $h_t$ as the state of the model at time $t$. 
We follow a dense connectivity pattern as~\cite{huang2017densely}, where each layer receives feature-maps from all preceding time-steps, generates a set of feature-maps for its own time-step and passes them on to all subsequent layers.
These layers are designed to be ``temporally causal'', in the sense that each layer has access only to past (backward) time steps, not forward steps.
%
%
After $T$ time steps, the final encoding is a concatenation of all the states $[h_1, \cdots h_T]$ which contains $K \times T$ feature maps in total.

%



\section{Evaluating Video Time} \label{sec:exp}
\subsection{Arrow of Time Prediction}
\label{sec:exp:task1}
\noindent\textbf{Datasets.}
For the first task we focus on two datasets. First, we use the dataset proposed in~\cite{pickup2014seeing} for evaluating arrow of time recognition, which we refer to as \emph{\textbf{Pickup-Arrow}}. The dataset contains 155 forward videos and 25 reverse videos, divided in three equal splits. We report the mean accuracy over all three splits. The second dataset is the more complex UCF101~\cite{soomro2012ucf101}, designed for action recognition. We ignore the action labels and use the videos in forward and backward order, both for training and test. We refer to this setup as \emph{\textbf{UCF101-Arrow}}.
We use split-1 of the dataset and again, report the mean accuracy.

\noindent\newline\textbf{Implementation details.} 
For C3D, we use the architecture proposed by~\cite{tran2015learning}, pretrained on Sports1M~\cite{KarpathyCVPR14}, and finetuned on UCF101. 
For LSTM and \ourmodel we use a VGG16~\cite{simonyan2014very} as the base network, pretrained on ImageNet~\cite{deng2009imagenet}, and then finetuned on UCF101.
The single-layer LSTM uses as input the $fc7$ activations with $512$ hidden units, while \ourmodel uses as input the last convolutional layer activations (before max-pooling), with the number of output feature maps in each step set to $K=12$.
The output passes through a final batch normalization, ReLU, max-pooling and fully connected layer with a $2$-dim output.
In terms of number of parameters, LSTM has 143M parameters (134M for the base VGG16 network, 9M for the LSTM units).
\ourmodel has 69M parameters (68M for the base VGG16 network, 1M for the time-step layers).
The difference in base network parameters between LSTM and \ourmodel is because LSTM ends with fully connected layers.
C3D has in total 78M parameters. 
%
%
%
All video encoders output a binary classification, indicating whether the video is played forward or backward.
The learning rate is set to $0.001$ for LSTM and ours and $0.0001$ for C3D as the learning was unstable with bigger learning rates.
%
%
Momentum and weight decay is set to $0.9$ and $5e-4$ for all the encoders.
All encoders receive the same $16$ frames as input.
During training we sample the frames from the whole video using a multinomial distribution, such that encoders see multiple sets of frames of a video.
During testing we sample frames from a uniform distribution in order to assure the same input is used for all the encoders.

\begin{figure}[t!]
\includegraphics[width=0.95\textwidth]{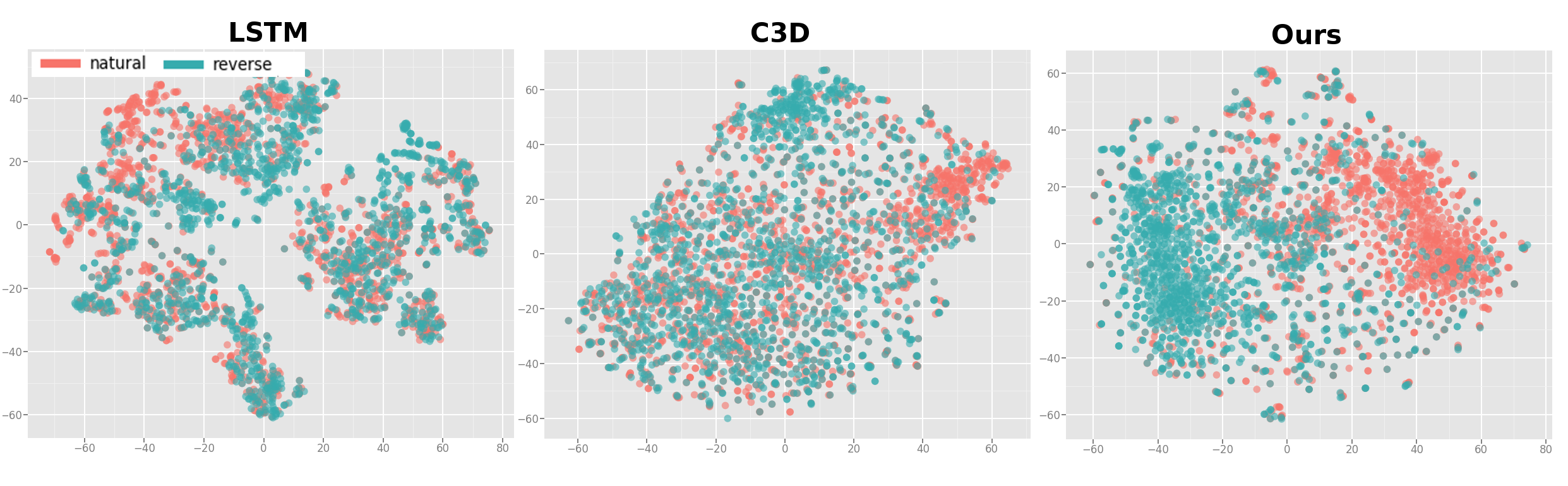}

\caption{t-SNE visualization of features from the last layer before the classification layer. Note the embeddings from our encoder are better clustered.}
\label{figure:tsne}
\end{figure}

\noindent\newline\textbf{Results.}
We report results in Table~\ref{table:task1} and plot in Fig.~\ref{figure:tsne} the t-SNE of all three model embeddings, collected from the last layer. On both datasets, the sequential encoder performs better than its hierarchical counterpart, indicating the hierarchical encoders are less good in distinguishing asymmetric temporal patterns. As explained in Section~\ref{sec:models}, C3D relies on learning correlations between input frames and, therefore, fails to capture conditional dependencies between ordered inputs.
From the t-SNE plots we observe that the embeddings from \ourmodel are better clustered, thus allowing for more accurate prediction of the arrow of time.
%
In the Pickup-Arrow dataset, LSTM is able to get close to the accuracy of \ourmodel, while in UCF101-Arrow the gap widens.
Despite the rapid progress in deep learning for video understanding, low-level visual features, hand-engineered for the task~\cite{pickup2014seeing}, with an SVM classifier are highly competitive. It appears there is much room for improvement of deep learning of video time representations.

\begin{SCfigure}
\caption{Accuracy of C3D, LSTM and ours in predicting the arrow of time for 5 best and worst performing classes in UCF101, sorted by C3D accuracies. Temporally causal actions like \emph{Billiards} appear to be easier than repetitive actions like \emph{Punch}. The class information is only used for illustration purposes.
}
\includegraphics[width=0.48\linewidth]{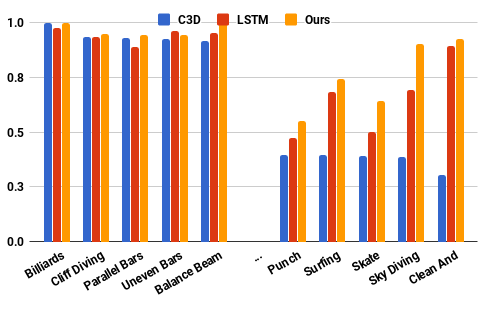}
\label{figure:task1_class}
\end{SCfigure}

It is noteworthy that the performance is not uniform across all videos.
When inspecting videos from particular UCF101 classes, see Fig.~\ref{figure:task1_class}, we observe that classes that one could claim are temporally causal (\eg, \emph{Billiards}, \emph{Still rings} and \emph{Cliff diving}) appear to be easier for all the methods.
That said, all encoders have difficulty for a big portion of the classes (79, 86 and 39 classes have less than $75\%$ accuracy for LSTM, C3D and \ourmodel respectively). Surprisingly, for a few classes, like \emph{Punch}, \emph{Drumming} and \emph{Military parade}, LSTM and C3D even report below chance accuracies. 

\begin{table}[t]
\caption{Arrow of time prediction results. Sequential encoders like LSTM and ours are better suited than C3D for this task. 
}
\label{table:task1}
\begin{sc}
\begin{adjustbox}{max width=\linewidth}
\resizebox{0.7\columnwidth}{!}{%
\begin{tabular}{lcc}
\toprule
 & UCF101-Arrow & Pickup-Arrow\\
\midrule
Chance          & 50.0 & 50.0 \\
LSTM  	    	& 67.5 & 80.0 \\
C3D  			& 57.1 & 57.1 \\
\ourmodel  		& \textbf{79.4} & \textbf{83.3} \\
\midrule
Pickup~\etal\cite{pickup2014seeing} & - & 80.6 \\
\bottomrule
\end{tabular}
}
\end{adjustbox}
\end{sc}
\end{table}
%

We conclude that even in ordinary videos, like UCF101, a temporal signal between successive video states exists.
Sequential video encoders including \ourmodel appear to be better for the task of the arrow of time, because they model the temporal conditional dependencies between visual states, instead of their correlations.

%

\subsection{Future Frame Selection}
\label{sec:exp:task2}
\noindent\textbf{Dataset.}
For the second task we focus on UCF101, because it is larger than the Pickup dataset. In UCF101 among others there exist action classes that are either static, like \emph{playing flute}, or periodic, like \emph{hula hoop}.
Clearly, in such videos temporal continuity is either trivial or purposeless, as in both cases the future frames will likely look nearly identical to (some of the) past frames. As having videos with a recognizable arrow of time also coincides well with videos where temporal continuity can be well evaluated, loosely following ~\cite{pickup2014seeing} we choose 24 classes with a distinguishable arrow of time.
We coin this dataset \emph{\textbf{UCF24-FutureFrame}}.
%



\noindent\newline\textbf{Implementation details.}
We train the models for this task using the same setup as Section~\ref{sec:exp:task1}.
During training, this task is formulated as a binary classification problem, where the encoder must predict whether a given frame is the correct future frame or not.
During testing, the encoder must select which frame is the correct future frame out of $C=5$ possible options. 
As a future frame we define the last frame $x_T$ of the video.
To increase the difficulty of the task, the incorrect frames are uniformly sampled from, $x_1, ..., x_{0.8T}$. 
To make sure the video encoders do not overfit to a specific temporal scale, we consider several $\tau$ during training and testing to define the last observed frames and report the mean accuracies\footnote{We pick $\tau$ from $ \{0.4,0.8,1.3,1.7,2.5,3.3,4.2,5.0,5.8,6.7,7.5,8.3\} \sec$ corresponding to the $5$-th, $10$-th, ... frame in the video with fps set to $12$ .}.
%

\noindent\newline\textbf{Results.}
We show the results in Table~\ref{table:task2}.
Future frame selection is harder than modeling the arrow of time, as it requires not only modeling of the arrow of time but also modeling temporal continuity.
We observe that C3D performs better than LSTM.
We attribute this to the fact that this task is spatio-temporal and C3D learns spatio-temporal filters, which can specialize un recognizing the most similar future frame.
%
\ourmodel outperforms both LSTM and C3D.
\ourmodel learns the temporal conditional dependencies between the spatial embeddings, instead of temporal correlations like C3D.
We also compare the encoders with a frame similarity baseline. 
We compute the cosine similarity between representations (normalized last convolutional layer of VGG16) of the last observed frame and the given future choices; the one with the highest score is considered as the correct answer.
All methods perform better than chance and the frame similarity baselines.
We conclude that for the temporal modeling task of future frame selection learning both spatial and temporal dependencies are necessary.


\begin{table}[t]
\caption{
Performance on future frame selection. Our model outperforms others due to the unshared convolutional filters which retain both spatial and temporal correlations.
}
\label{table:task2}
\begin{sc}
\begin{adjustbox}{max width=\linewidth}
\resizebox{0.6\columnwidth}{!}{%
\begin{tabular}{lc}
\toprule
 & UCF24-FutureFrame \\
\midrule
Chance        & 25.0 \\
LSTM  	      & 46.9 \\
C3D  		  & 51.9 \\
\ourmodel	  & \textbf{56.3} \\
\midrule
Frame similarity & 24.2 \\
\bottomrule
\end{tabular}
}
\end{adjustbox}
\end{sc}
\end{table}
%

\subsection{Action Template Classification}
\label{sec:exp:task3}
\noindent\textbf{Dataset.} For the third task we focus on the Something-Something dataset \cite{goyal2017something}, where the action class labels are template-based textual descriptions like \emph{Dropping [something] into [something]}. The Something-Something dataset is crowd-sourced and contains $86,017$ training and $11,522$ validation videos of length $2-6 \sec$, across $174$ action categories in total. The action classes represent physical actions instead of high-level action semantics (see Fig.~\ref{fig:tasks}), such as \emph{Pouring something into something} or \emph{Pouring something until it overflows}. As classes cannot be recognized by indirect cues like object or scene type, the temporal reasoning plays an important role.

\noindent\newline\textbf{Implementation details.}
For this experiment we follow the setup proposed by~\cite{wang2016temporal}.
As base network we choose Inception with Batch Normalization (BN-Inception), pretrained on ImageNet. 
The input feature for LSTM are the activations from the global averaging pooling layer. Since \ourmodel accepts convolutional feature maps, we use the activations from the layer before average pooling, after adding an extra convolutional layer to reduce the number of feature maps from $1024$ to $256$ for better efficiency.
We set the number of time-steps to $4$ for all the models, except for C3D, which is pretrained with $16$ frames. For all models, but C3D, we use a fully connected layer with $512$ units before the classification layer. 

\noindent\newline\textbf{Results.}
We show results in Table~\ref{table:task3}. Similar to the second task, LSTM cannot easily learn the temporal conditional dependencies required for the recognition of the physical and time-specific actions.
C3D is better than LSTM. Again, \ourmodel outperforms both C3D and LSTM. 
\ourmodel also outperforms the Temporal Segmentation Network \cite{wang2016temporal} (TSN) by a significant margin.
The reason is after splitting a video into segments, TSN discards temporal structure by aggregating video segment representations via average pooling.
%
We also compare with the concurrent work of Zhou~\etal~\cite{zhou2017temporal}, reporting $29.8$ and $58.2$ prec@1 and prec@5 respectively, slightly less than \ourmodel for this setting. Their model also aims at learning temporal relations between ordered frames (sets of two frames, three frames, etc.) via fully connected layers, and thus the model cannot be easily adapted for higher order frame relations.


We conclude that for template-based action classification, modelling temporal conditional dependencies is important.
Similar to the second task, the model should be able to parameterize the temporal conditional dependencies per time step freely.
If not, sharing parameters through time leads to worse results than not modeling the dependencies at all.


\begin{table}[t]
\caption{Performance on Task III. C3D and ours are able to learn the temporal dependencies while LSTM can not. 
%
%
Temporal modeling is crucial for this task.}
\label{table:task3}
\begin{sc}
\begin{adjustbox}{max width=\linewidth}
\resizebox{0.6\columnwidth}{!}{%
\begin{tabular}{lcccc}
\toprule
 & \multicolumn{2}{c}{Something-Something}\\
\cmidrule(lr){2-3}
 & Prec@1 & Prec@5 \\
 \midrule
Chance      & 0.5 & 3.0\\
LSTM  	    & 15.7 & 39.5\\
C3D			& 28.2 & 56.3 \\
\ourmodel	& \textbf{30.4} & \textbf{59.3}\\
\midrule
Wang~\etal~\cite{wang2016temporal} 			& 16.0  & 41.1\\
\bottomrule
\end{tabular}
}
\end{adjustbox}
\end{sc}
\end{table}

\section{Conclusions}
In this work, we investigated and evaluated the manifestation of time in video data. Our meta analysis quantifies video time with respect to temporal asymmetry, continuity and causality by three corresponding tasks. Moreover, we proposed a new video time encoder and provided in-depth analysis of LSTM, C3D and our proposed encoder in this setting. We observed LSTM is better than C3D in handling our temporal asymmetry task. As the need for joint modeling of spatio-temporal data increases in our tasks measuring temporal continuity and causality, C3D is outperforming LSTM. Our proposed \ourmodel consistently outperforms both C3D and LSTM on all three tasks. An important factor appears to be the unshared parameterization of our proposed encoder in modeling temporal transitions. We believe that more detailed understanding of time holds promise for the next iteration of encoders. 
%
%


\bibliography{egbib}
\end{document}